\documentclass[letterpaper, 10 pt, conference]{ieeeconf}  

\IEEEoverridecommandlockouts                              

\usepackage{graphics} 
\usepackage{epsfig} 
\usepackage{times} 
\usepackage{amsmath} 
\usepackage{amssymb}  

\usepackage{booktabs, multirow} 
\usepackage{float}

\usepackage{algorithm}
\usepackage{algpseudocode}
\usepackage[table]{xcolor}

\usepackage{caption}
\usepackage{pifont}
\newcommand{\cmark}{\ding{51}}%
\newcommand{\xmark}{\ding{55}}%

\usepackage{xcolor}

\title{\LARGE \bf
Performance-Guided Refinement for Visual Aerial Navigation \\ using Editable Gaussian Splatting in FalconGym 2.0
}

\author{Yan Miao$^{1}$, Ege Yuceel$^{1}$, Georgios Fainekos$^{2}$, Bardh Hoxha$^{2}$, Hideki Okamoto$^{2}$ and Sayan Mitra$^{1}$
\thanks{$^{1}$University of Illinois at Urbana-Champaign}
\thanks{$^{2}$Toyota Research Institute of North America}
}

\begin{document}

\maketitle
\thispagestyle{empty}
\pagestyle{empty}

\begin{abstract}
Visual policy design is crucial for aerial navigation. 
However, state-of-the-art visual policies often overfit to a single track and their performance degrades when track geometry changes. 
We develop \emph{FalconGym~2.0}, a photorealistic simulation framework built on Gaussian Splatting (GSplat) with an \emph{Edit API} that programmatically generates diverse static and dynamic tracks in milliseconds. 
Leveraging FalconGym 2.0's editability, we propose a \emph{Performance-Guided Refinement (PGR)} algorithm, which concentrates visual policy's training on challenging tracks while iteratively improving its performance.
Across two case studies (fixed-wing UAVs and quadrotors) with distinct dynamics and environments, we show that a single visual policy trained with PGR in FalconGym~2.0 outperforms state-of-the-art baselines in generalization and robustness: it generalizes to three unseen tracks with 100\% success \emph{without} per-track retraining and maintains higher success rates under gate-pose perturbations. 
Finally, we demonstrate that the visual policy trained with PGR in FalconGym 2.0 can be \emph{zero-shot sim-to-real transferred} to a quadrotor hardware, achieving a 98.6\% success rate (69 / 70 gates) over 30 trials spanning two three-gate tracks and a moving-gate track.
\end{abstract}

\section{Introduction}

Visual aerial navigation is critical for applications such as mapping, search-and-rescue, environmental monitoring and racing. 
Recent progress in photorealistic simulation environments has fueled zero-shot sim-to-real success for visual aerial navigation.
Notably,
SOUS VIDE \cite{low2024sousvide} used Gaussian Splatting (GSplat) \cite{kerbl3Dgaussians} to reconstruct an indoor lab and achieved zero-shot sim-to-real navigation;
FalconGym~\cite{miao2025zeroshotsimtorealvisualquadrotor} used NeRF~\cite{10.1145/3503250} to build digital twins of three racing tracks and demonstrated zero-shot sim-to-real transfer of quadrotor gate crossing via imitation learning; 
Geles \emph{et~al.}~\cite{DBLP:conf/rss/GelesBRX024} reported strong sim-to-real performance with a vision-only asynchronous actor-critic on three quadrotor racing tracks. 
While effective, these benchmarks face generalization limits:
all three achieve high performance on their training tracks but do not generalize to unseen tracks (as we also confirm in Section \ref{sec:experiments}), restricting broader applicability. 
GRaD-Nav \cite{chen2025gradnavefficientlylearningvisual} improves robustness to different gate positions and background distractors by using one  policy, but still relies on constructing multiple GSplat tracks first and training in those.

\begin{figure}[htbp]
    \centering
    \includegraphics[width=\linewidth]{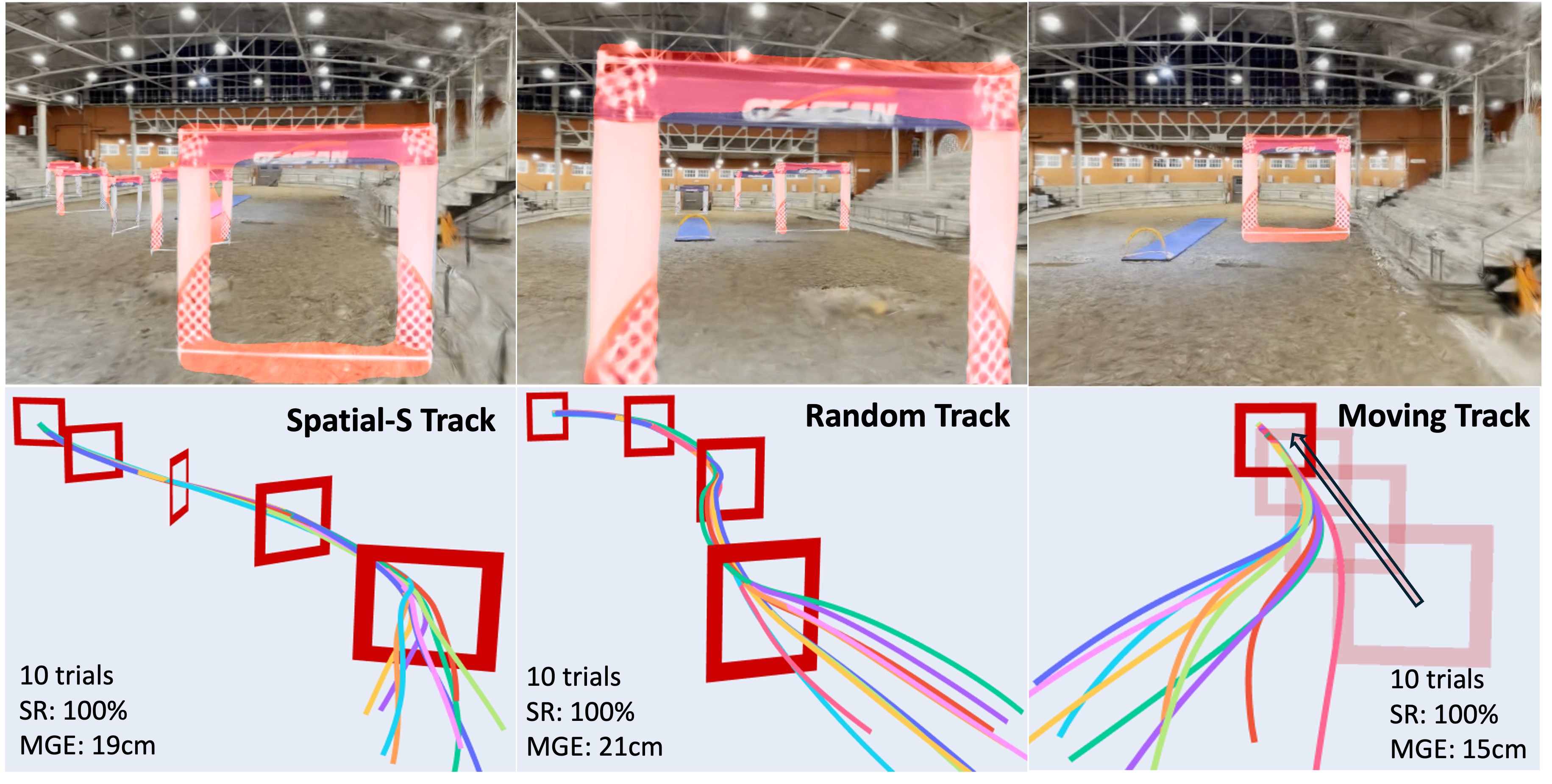}
    \caption{\small{
    \textbf{Trajectories for UAV case study in FalconGym 2.0 across three unseen tracks (Spatial-S, Random and Moving).} 
    Top row: red overlays visualize predicted gate masks (Section \ref{sec:vision-controller-architecture}). Bottom row: 10 trials per track from different initial states; the translucent gates in the Moving track show the gate's past positions.
    }}
    \label{fig:uav_trajectory_plot}
\end{figure}

In this paper, we focus on developing a single visual policy that can traverse a family of tracks, where each track consists of an ordered sequence of tight racing gates, as shown in Figure \ref{fig:uav_trajectory_plot} and Figure \ref{fig:quadrotor_trajectory_plot}. 
To enable such cross-track generalization from limited real data, we develop \emph{FalconGym~2.0}. 
From just minutes of real-world videos, FalconGym~2.0 can generate arbitrarily many synthetic yet photorealistic tracks for training without additional real-world data collection. 
This is achieved by our \emph{Edit API}, as shown in Figure \ref{fig:editable-gsplat}, that allows programmatic editing of gates in the GSplat to create diverse tracks in milliseconds. 
Beyond static edits, our Edit API also supports 4D time-varying simulation with moving gates, enabling dynamic tracks in Section \ref{sec:experiments}. 
Although this paper focuses on aerial navigation and gate-editing, the same Edit API could readily extend to broader robotics settings (e.g., obstacle placement for ground robots).

FalconGym 2.0's editable capability to generate arbitrarily many photorealistic tracks unlocks a range of visual policy training strategy from active learning to curriculum learning ~\cite{soviany2022curriculumlearningsurvey}.
We develop a \emph{Performance-Guided Refinement (PGR)} algorithm that: (i) identifies challenging tracks where the visual policy underperforms; (ii) synthesizes similar yet slightly different challenging tracks with the Edit API to augment the training dataset; and (iii) iteratively refines the visual policy via imitation learning from a state-based expert.

Beyond the Edit API and the PGR algorithm it enables, FalconGym~2.0 improves over FalconGym  \cite{miao2025zeroshotsimtorealvisualquadrotor} via: 
(i) replacing NeRF rendering with fast GSplat to accelerate training; 
(ii) eliminating expensive motion capture with an accessible ArUco marker for world-frame simulation reconstruction. 

Through two case studies (fixed-wing UAVs and quadrotors) with different dynamics, environments and gate geometries, we show that our visual policy trained with PGR can generalize to three unseen tracks in FalconGym 2.0 with 100\% success. 
Moreover, we outperform state-of-the-art visual baselines~\cite{miao2025zeroshotsimtorealvisualquadrotor,DBLP:conf/rss/GelesBRX024} in both generalization and robustness: our single visual policy operates across three unseen tracks, where baselines require separate per-track policies, and we maintain higher performance under gate-pose perturbations.
Finally, the visual policy trained with PGR in FalconGym 2.0 can zero-shot transfer to a quadrotor hardware, achieving 98.6\% (69 / 70 gates) success rate in 30 hardware trials spanning two 3-gate tracks and a moving-gate track, as shown in Figure \ref{fig:quadrotor_hardware_trajectory_plot}.

In summary, our contributions are:
(1) \emph{FalconGym 2.0}: an editable photorealistic simulation framework based on GSplat that supports fast world-frame modification of environment configurations. 
(2) \emph{Performance-Guided Refinement (PGR)}: an algorithm that leverages FalconGym 2.0's editability to expose the visual policy's training to challenging tracks and iteratively improve performance. 
(3) \emph{Zero-Shot Sim-to-real}: Our visual policy trained with PGR in FalconGym 2.0 can be zero-shot transferred to a real quadrotor hardware to traverse three unseen tracks.

\section{Related Work}

\paragraph{Sim-to-real in robotics.}
Sim-to-real transfer is a longstanding goal in robotics, due to its efficient and safe training before deployment. 
With advances in photorealistic scene reconstruction like NeRF~\cite{10.1145/3503250} and 3D Gaussian Splatting (GSplat)~\cite{kerbl3Dgaussians}, recent robotics work increasingly trains in high-fidelity simulators. 
NeRF2Real~\cite{Byravan2022NeRF2RealST} and RialTo~\cite{torne2024rialto} demonstrate NeRF-to-real transfer for humanoid navigation and robot manipulation, respectively.
Vid2Sim~\cite{xie2024vid2sim} converts real-world videos into interactive simulators for urban navigation using GSplat reconstruction, while RoboSplat~\cite{robosplat} and Splat-MOVER \cite{shorinwa2024splat} leverages GSplat for domain randomization to  improve robot manipulation performance. 

In aerial navigation, FalconGym~\cite{miao2025zeroshotsimtorealvisualquadrotor} constructs NeRF-based digital twins of racing tracks and achieves zero-shot sim-to-real quadrotor gate crossing via imitation learning, while SOUS VIDE~\cite{low2024sousvide} uses GSplat to build an indoor digital twin and also reports zero-shot transfer. 
Geles \emph{et~al.}~\cite{DBLP:conf/rss/GelesBRX024} train a vision-only policy using async actor-critic and demonstrate high-speed racing ($40\,\mathrm{km/h}$) across three tracks.
Despite strong sim-to-real results, these benchmarks generally require per-track training and struggle to generalize to unseen tracks. 

\paragraph{Policy Refinement by Environment Shaping}
Curriculum-based environment shaping \cite{soviany2022curriculumlearningsurvey} has been used to improve control policy in robotics by adaptively shaping training environments.
It has been successful in simulation for RL agents in bipedal walkers \cite{pmlr-v100-portelas20a}.
\cite{wang2025environmentpolicylearningrace} uses a similar ``environment policy'' idea for quadrotor navigation. 
However, these work used state-based control and did not involve rendering or visual policy, while our PGR algorithm adapts an editable photorealistic GSplat environment to improve visual policy performance for aerial navigation.

\paragraph{Editable Gaussian Splatting}
Recent work focuses on using editable GSplat to generate synthetic but photorealistic environment.
GaussianEditor~\cite{chen2023gaussianeditorswiftcontrollable3d} enables text-guided appearance and geometry edits of objects in GSplat.
VCEDIT \cite{wang2025viewconsistent3deditinggaussian} ensures multi-view consistency when using diffusion-guided GSplat edits. 
Instruct-GS2GS \cite{igs2gs} provides instruction-based editing of GSplat scenes via 2D diffusion.
%
%
While these work advance GSplats scene editing in computer vision, they focus on improving photorealism and accuracy, rather than connecting to downstream closed-loop robotics applications.
In contrast, FalconGym~2.0 not only provides open-source Edit APIs to edit objects in GSplat, but also couples this editability with a performance-guided refinement algorithm that iteratively improves downstream visual policy's performance in the closed loop. 

%


\section{Methods}
\label{sec:methods}

\begin{figure}[htbp]
    \centering
    \includegraphics[width=\linewidth]{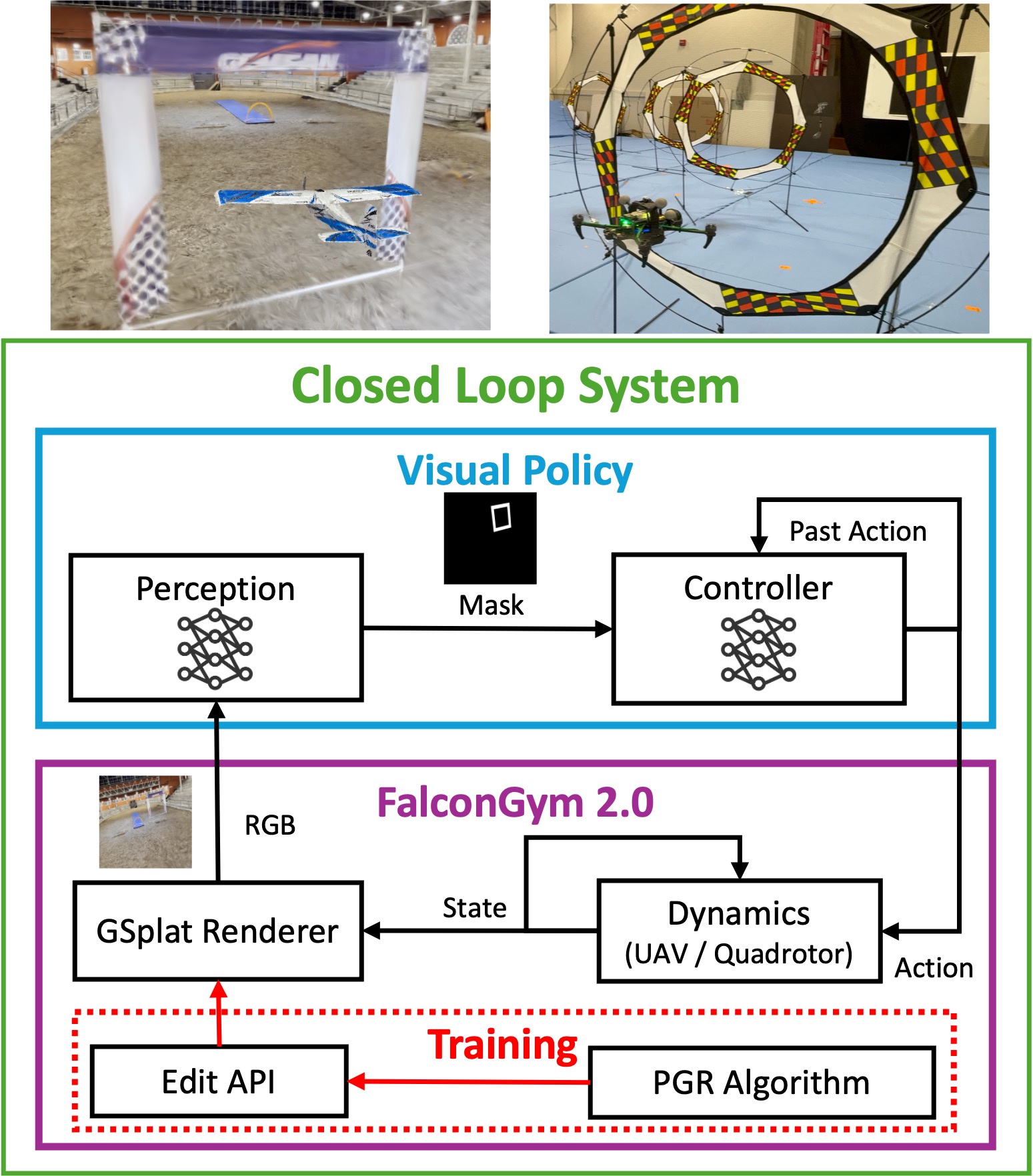}
    \caption{\small{
    \textbf{Closed-loop system in FalconGym 2.0: }
    we provide dynamics for a fixed-wing UAV and a quadrotor, and a GSplat renderer that produces photorealistic RGB from arbitrary camera poses in either scene. 
    At each timestep, the dynamics propagate the state, the renderer generates an RGB image, a perception module predicts a gate mask, and a controller consumes the mask plus past actions to predict the next action. 
    During training, a \emph{Performance-Guided Refinement (PGR)} algorithm (Section \ref{sec:min-max-optimization}) focuses training on challenging tracks generated using \emph{Edit API} (Section \ref{sec:editable-gsplat})
    }}
    \label{fig:closed-loop}
\end{figure}

\begin{figure*}[htbp]
    \centering
    \includegraphics[width=\linewidth]{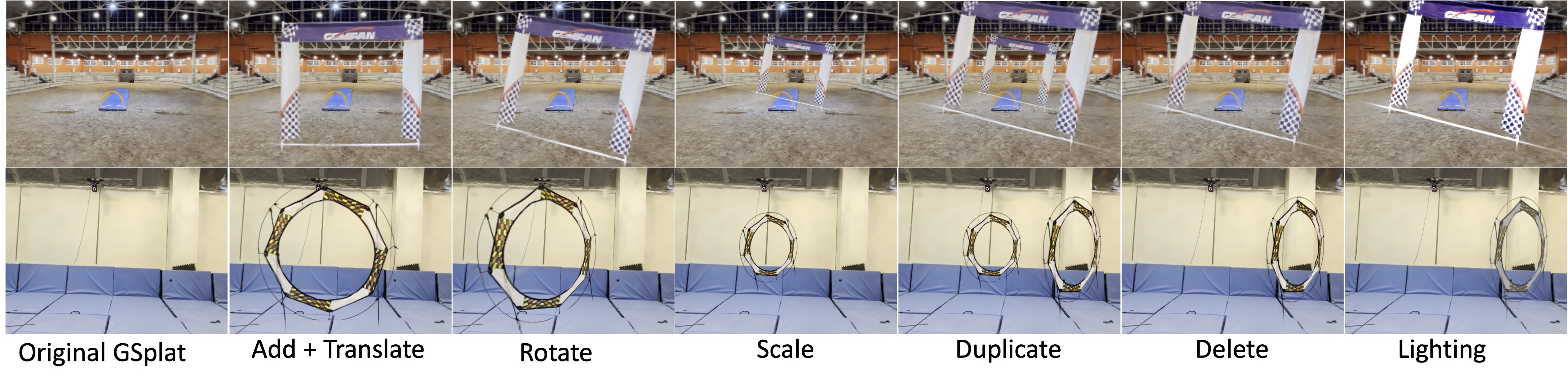}
    \caption{\small{
    \textbf{Edit API in FalconGym~2.0.} Our Edit API (Section \ref{sec:editable-gsplat}) provides world-frame programmatic placement of objects while the backend handles all coordinates and camera-to-world transform. The seven API: \texttt{add}, \texttt{translate}, \texttt{rotate}, \texttt{scale}, \texttt{duplicate}, \texttt{delete}, and \texttt{lighting}, allow users to modify object pose, size, and appearance to generate a photorealistic 4D simulation environment. 
    Shown are gates edits across two environments. This editable capability enables PGR algorithm to improve visual policy (Section \ref{sec:min-max-optimization}).
    }}
    \label{fig:editable-gsplat}
\end{figure*}

In this paper, we consider visual aerial navigation in the gate-based track setting, where each track is an ordered sequence of tight racing gates and the aerial vehicle must safely traverse a set of tracks using only visual feedback. 
Successful trajectories are shown in FalconGym~2.0 for a fixed-wing UAV (Figure \ref{fig:uav_trajectory_plot}) and a quadrotor (Figure \ref{fig:quadrotor_trajectory_plot}), and on hardware for the quadrotor (Figure \ref{fig:quadrotor_hardware_trajectory_plot}).

Prior work~\cite{miao2025zeroshotsimtorealvisualquadrotor,DBLP:conf/rss/GelesBRX024} tackles this same aerial navigation problem and reports strong gate-crossing performance; however, these methods require \emph{per-track} training and a visual policy trained on one track does not generalize reliably to unseen tracks. 
This overfitting could be costly in practice, since recollecting data and retraining for every new track is time-consuming. 
Therefore, our goal is to develop a single visual policy that operates zero-shot on unseen tracks, eliminating per-track retraining while maintaining robustness to track-configuration changes. 
In this paper, we restrict the tracks to be dynamically feasible and observable (when the aerial vehicle crosses current gate center orthogonally, the next gate must be within the camera’s field of view). 
This observability requirement that keeps the next gate in sight is reasonable in a visual-only setting without knowledge of the map.

Our method comprises three components. 
First, we present \emph{FalconGym 2.0} (Sec.\ref{sec:editable-gsplat}), a GSplat-based photorealistic simulation framework coupled with an \emph{Edit API} that enables dynamic gate placement, improving on ~\cite{low2024sousvide, miao2025zeroshotsimtorealvisualquadrotor} static simulation.
Second, we design a visual policy (Section \ref{sec:vision-controller-architecture}) in the closed loop in FalconGym 2.0 that mitigates the overfitting observed in earlier work \cite{miao2025zeroshotsimtorealvisualquadrotor} through modular architecture that separates perception and control.
Third, we introduce a novel \emph{Performance-Guided Refinement (PGR)} algorithm (Section \ref{sec:min-max-optimization}) that uses FalconGym 2.0's editability to expose visual policy to challenging tracks and iteratively refine its performance.

\subsection{FalconGym 2.0: Editable GSplat}
\label{sec:editable-gsplat}

We propose a photorealistic simulation framework \emph{FalconGym 2.0}, that is capable of developing and testing different visual policies, while improving on FalconGym ~\cite{miao2025zeroshotsimtorealvisualquadrotor} by replacing NeRF with GSplat for fast rendering; substituting motion-capture with an ArUco marker for world-frame alignment. \footnote{
More specifically, to construct FalconGym~2.0, a human operator uses the onboard camera to capture a 3-minute video across the flying arena from diverse viewpoints.
We recover camera intrinsics and initial poses with COLMAP~\cite{7780814}.
Because COLMAP’s frame is arbitrary, we align it to a physically meaningful world frame by placing an easily accessible ArUco marker in the scene, thereby eliminating the need for expensive motion capture.
Using OpenCV’s ArUco detector, we locate the marker center in a subset of images where the marker is visible and treat it as the global origin.
We then compute the rigid transform between the COLMAP and world frames via the Kabsch-Umeyama method~\cite{Kabsch:a15629}.
Finally, the images and poses are fed to the NeRFStudio \texttt{Splatfacto} pipeline~\cite{nerfstudio} to train a photorealistic GSplat scene in world coordinates.
}
More importantly, FalconGym~2.0 introduces an \emph{Edit API} that's capable of generating arbitrarily many training tracks.

%

\paragraph{GSplat Scene representation}
A trained GSplat scene consists of a set of $N$ anisotropic Gaussians, i.e.
$\mathcal{S}=\big\{ \, (\mu_j,\Sigma_j,c_j,\alpha_j) \, \big\}_{j=1}^N,$
where $\mu_j \in \mathbb{R}^3$ is the mean; $\Sigma_j \in \mathbb{R}^{3\times3}$ is the covariance, parameterized via a rotation $R_j \in \mathrm{SO}(3)$ and per-axis scales $s_j \in \mathbb{R}_{>0}^3$ (i.e., $\Sigma_j = R_j \operatorname{diag}(s_j^2) R_j^\top$); $c_j \in \mathbb{R}^3$ is the color; and $\alpha_j \in [0,1]$ is the opacity. 
For rendering an image as seen by a camera in this scene $S$, the $N$ 3D Gaussians are projected onto the camera's image plane then their colors are blended according to their depths relative to the camera.

\paragraph{Edit API}
Because NeRFStudio \cite{nerfstudio} (the library we used to train GSplat) optimizes scenes in an internal coordinate frame, programmatic editing is inconvenient for users who reason in a world frame.
FalconGym~2.0 instead exposes world-frame edits and handles all coordinate conversions and camera-to-world transformations in the backend.
To edit a scene in FalconGym~2.0, users can first select an object either via predefined Gaussian IDs (e.g., gate primitives from another GSplat scene provided by us) or via user-defined world-frame bounding boxes, then apply pose, color and scale edits via our API.
Concretely, we provide seven composable operations, as shown in Figure \ref{fig:editable-gsplat}:
\texttt{add()} inserts objects from another GSplat scene;
\texttt{translate()} moves selected objects to user-defined poses;
\texttt{rotate()} rotates the selected objects around its center for user-defined rotation;
\texttt{scale()} scales the selected objects' sizes;
\texttt{duplicate()} clones the selected object;
\texttt{delete()} removes selected objects;
\texttt{lighting()} adjusts gaussian-level colors for selected objects.
Each operation updates the selected objects' corresponding Gaussians’ $\mu_j$, $\Sigma_j$ (via $R_j,s_j$), $c_j$, and $\alpha_j$ on the backend. 
Since each API is a tensor operation, on average each operation finishes $\sim 0.004$s with a 4090 GPU.
Users can freely combine any of the 7 API to accomplish complex edits and build customized 4D simulations. (e.g., moving gates in Section~\ref{sec:experiments}). 
Although our experiments focus on aerial navigation with gates, we expect the same Edit API to work on broad robotics tasks that would benefit from photorealistic editable GSplat scenes.


\subsection{Visual Policy in the Closed Loop}
\label{sec:vision-controller-architecture}

In this subsection, we introduce our closed-loop system architecture and visual policy design, as shown in Figure \ref{fig:closed-loop}.
The previous subsection has introduced the Edit API for moving objects (i.e. gates) within GSplat scenes; here we focus on moving the aerial vehicle (i.e., the onboard camera).
Although the aerial vehicle could be placed in arbitrary poses in FalconGym 2.0, closed-loop control requires its motion obey physically meaningful dynamics.
FalconGym 2.0 provides two aerial dynamics models: a Dubins airplane dynamics model for fixed-wing UAVs \cite{Owen2015} and a quadrotor dynamics model \cite{Sabatino2015QuadrotorCM}. 
The dynamics model is implemented in a plug-and-play fashion so users can substitute it with alternative platform's dynamics (e.g., VTOL or ground robots).

\begin{figure}[htbp]
    \centering
    \includegraphics[width=\linewidth]{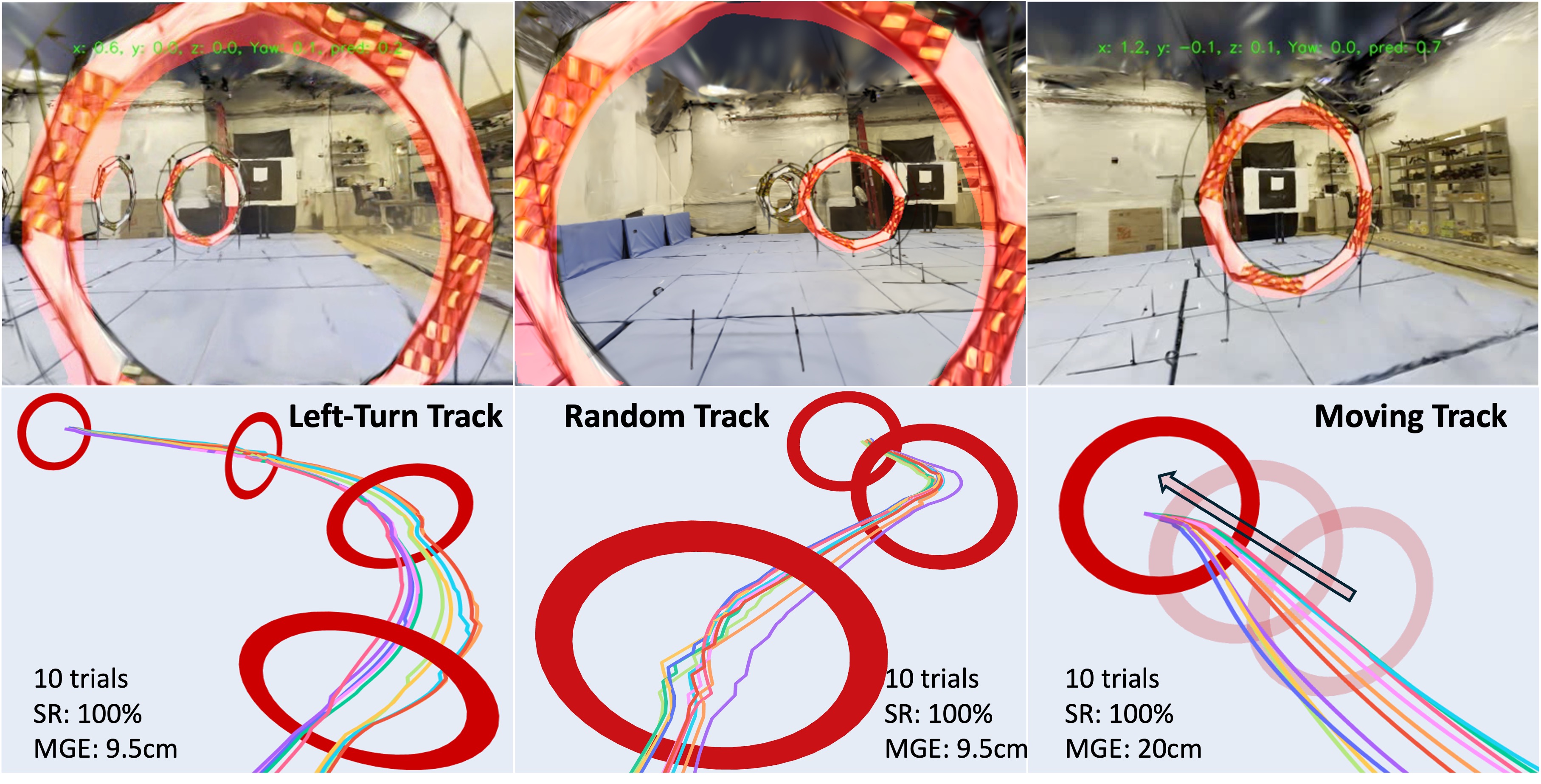}
    \caption{\small{
    Trajectories for quadrotor case study in FalconGym 2.0
    across three unseen tracks (Left-Turn, Random and Moving).
    }}
    \label{fig:quadrotor_trajectory_plot}
\end{figure}

With renderer and dynamics in place, next we focus on the visual policy design.
\cite{miao2025zeroshotsimtorealvisualquadrotor} is quite successful in zero-shot sim-to-real quadrotor navigation by an end-to-end ViT policy.
However, as also acknowledged by their authors, their visual policy trained on one track tends to specialize to that track.
Our further experiments confirm \cite{miao2025zeroshotsimtorealvisualquadrotor}'s overfitting: even after removing the gates entirely using our Edit API, \cite{miao2025zeroshotsimtorealvisualquadrotor}'s visual policy continued to fly around the memorized route, suggesting reliance on background appearance rather than gate-relevant features.
Moreover, \cite{miao2025zeroshotsimtorealvisualquadrotor} has to run the hardware experiments offboard due to a heavy dual-Vision-Transformer (ViT) policy, which also required ground-truth state of the next gate.

Therefore, to mitigate overfitting and reduce model complexity for onboard deployment, we design a modular architecture that decouples perception from control, 
inspired by \cite{DBLP:conf/rss/GelesBRX024} which trains a RL algorithm based on gate-perception masks to fly through racing gates.
First, we design a perception module that predicts a gate mask.
Next, a lightweight controller consumes this mask with a short history of past controls to output the next action, as shown in Figure \ref{fig:closed-loop}.
As validated in Section \ref{sec:experiments}, this modular pipeline not only improves generalization, but also supports on-board execution due to the smaller model size compared to \cite{miao2025zeroshotsimtorealvisualquadrotor}'s dual-ViT setup.

\paragraph{Gate-Detection Perception Module}
We train the perception module completely in FalconGym 2.0.
As shown in the blue box in Figure \ref{fig:closed-loop}, the perception module takes an onboard RGB image and predicts a binary mask where white pixels indicate gates and black pixels indicate background. 
We adopt a U-Net~\cite{10.1007/978-3-319-24574-4_28} backbone for gate segmentation for its strong performance on dense prediction.
To obtain ground-truth masks, we automatically generate them analytically via standard 3D-to-2D projection techniques in computer vision. 
Because gate positions (placed in FalconGym 2.0 through our Edit API) are known and the geometry (e.g. diameter) is measured, we can project 3D gates to a 2D image plane using known camera matrix.
Pixels whose coordinates lie within the gate's projected ring (i.e., between inner and outer boundaries) are labeled white, and all others black.


To collect training data for the U-Net, we: (i) leverage the Edit API to place a two-gate track in the workspace; (ii) sample a feasible camera pose at an appropriate distance with yaw roughly oriented toward the track; and (iii) render the RGB image and compute its ground truth 2D mask as above. 
We sample two-gate tracks because we want the U-Net to learn about scenarios where one gate might be occluded by the other, and experiments show our U-Net could generalize to multi-gate detection.
We gather RGB images (square-gate UMX for UAV case study and circular-gate for quadrotor case study) and train separate U-Net for each case study with supervised learning on these image-mask pairs. Qualitative results of the U-Net are shown in Figure \ref{fig:uav_trajectory_plot} and \ref{fig:quadrotor_trajectory_plot} in FalconGym 2.0, while Figure \ref{fig:quadrotor_hardware_trajectory_plot} visualizes its performance when zero-shot deployed on hardware.

\paragraph{Controller}
With a predicted mask, the controller receives an explicit geometry-focused signal of gate locations, potentially reducing reliance on background cues and mitigating overfitting. 
Next, we further improve the architecture of~\cite{miao2025zeroshotsimtorealvisualquadrotor} by removing IMU inputs and instead feeding a short history of past controls to the controller (blue box in Figure \ref{fig:closed-loop}), which provides implicit temporal context.
Next, the controller training follows a similar imitation learning procedure as in \cite{miao2025zeroshotsimtorealvisualquadrotor}: we first implement a state-based expert that flies through different tracks in simulation; at each timestep, we render the onboard RGB image and record the state-based controller's expert action.
The RGB image is passed through the trained U-Net to obtain a binary mask, and we form supervised pairs where the masked image coupled with the past control actions are used to predict the current action to train the controller.

Thanks to the Edit API, now we can synthesize essentially arbitrarily many tracks in FalconGym 2.0 to train both perception and controller without additional per-track real-world effort required by~\cite{low2024sousvide, miao2025zeroshotsimtorealvisualquadrotor, DBLP:conf/rss/GelesBRX024}. 
To sample efficiently, our unique design choice is to train on \emph{two-gate tracks}. 
Intuitively, the initial state together with two successive gates spans the local geometric variability of longer courses; a controller that performs well on such segments could generalize well to multi-gate tracks by invariance and composition, as is empirically confirmed in Section \ref{sec:experiments}.


\subsection{Performance-Guided Refinement Training}
\label{sec:min-max-optimization}

A straightforward method to collect training data for the visual policy would be to uniformly sample the two-gate track space that is dynamically feasible and observable (as defined at the start of this section).
%
%
However, uniform sampling can be sample-inefficient in a large high-dimensional workspace. 
With our Edit API, we can steer training data collection toward the visual policy's weak spots and iteratively refine to improve the visual policy. 

Inspired by adversarial training \cite{10.1145/3422622} and min-max problem \cite{Shimizu1997} which tries to minimize the possible loss for a worst case (maximum loss) scenario, we cast our aerial navigation as a min-max problem:
\begin{equation}
\min_{\theta} \; \max_{g \in G} \; \mathbb{E}_{\tau \sim \pi_\theta} \big[\,\mathcal{L}(\tau; g)\,\big],
\label{eq:minmax-train}
\end{equation}
where $g$ parameterizes a two-gate track, $G$ is the dynamically feasible and observable two-gate space, and $\mathcal{L}$ measures task performance on a trajectory rollout $\tau$ achieved by the visual policy (perception and controller) $\pi$ parameterized by $\theta$. We define the task performance function as $
\mathcal{L}(\tau; g) 
= \mathbf{1}\{\text{collision or timeout}\}
+ \lambda_{\mathrm{pos}} \,\big\|\, p' - c(g) \,\big\|_2 ,
$
where $c(g) \in \mathbb{R}^3$ is the gate center, and $p' \in \mathbb{R}^3$ is the aerial vehicle's position when crossing the target gate plane. 
The second term is only evaluated on successful crossings.
%
%
Intuitively, “maximizer” proposes challenging gate placements to cause visual policy $\pi_\theta$'s poor performance, while “minimizer” updates $\pi_\theta$ to reduce loss to improve performance. 
This way, training is focused on where visual policy  under-performs.

\begin{algorithm}[htbp]
\caption{\small{Performance-Guided Refinement of Visual Policy}}
\label{alg:minmax}
\begin{algorithmic}[1]
\Require Gate space $G$ partitioned into $M$ grids, state‐based expert policy $\pi_s$, iterations $T$, validation gate set $G_{\mathrm{val}}$
\Ensure Trained visual policy $\pi_\theta$
\State $\mathcal{D} \gets \varnothing$ \Comment{Training dataset}
\State Sample initial tracks $G_{1}\subset G$ uniformly at random
\For{$t = 1$ to $T$}
  \ForAll{two-gate track $g \in G_t$}
    \State Run closed‐loop trajectory rollout with $\pi_s$ and $g$
    \State Collect trajectory (action, image) and add to $\mathcal{D}$
  \EndFor
  \State Train visual policy $\pi_\theta$ on $\mathcal{D}$
    \For{$i=1$ \textbf{to} $M$}
        \Comment{\small{evaluate grid-wise performance}}
      \State $\ell_i \gets \frac{1}{|G_{\mathrm{val}}\cap M_i|}\sum_{g\in G_{\mathrm{val}}\cap M_i} \mathcal{L}\bigl(\pi_\theta;g\bigr)$
    \EndFor
    \State Compute normalized grid weights:
    \State $
      w_i \; \gets \frac{\ell_i}{\sum_{i=1}^M \ell_i}
      \quad\forall i=1,\dots,M
    $
    \State $w_i \leftarrow (1-\beta)\,w_i + \frac{\beta}{M}$ \Comment{Avoid Mode Collapse}
    \State Generate next set $G_{t+1}$ by: first choosing $M_i$ with probability $w_i$ and then sample $g \sim M_i$ uniformly
\EndFor
\end{algorithmic}
\end{algorithm}

Yet directly solving Equation~\eqref{eq:minmax-train} is infeasible, so we implement grid-based \emph{Performance-Guided Refinement (PGR)}, as shown in Algorithm~\ref{alg:minmax}.
We partition the two-gate space $G$ into $M$ grids $\{M_i\}_{i=1}^M$.
During the first iteration, we synthesize $G_1$ by first uniformly sampling the grids and then drawing two-gate layouts uniformly within selected grids.
We delete the dynamically infeasible and unobservable tracks automatically using the expert state-based controller and geometric calculation.
Next, for each iteration $t$: we (i) collect trajectory rollouts on a batch of two-gate layouts $G_t$ using the expert state-based controller to augment the perception and imitation dataset $\mathcal{D}$, (ii) train the visual policy $\pi_\theta$ on $\mathcal{D}$, (iii) evaluate per-grid validation performance $\ell_i$ using a held-out set $G_{\mathrm{val}}$, and (iv) \emph{resample} the next batch $G_{t+1}$ by first drawing grids with probability proportional to $\ell_i$ and then sampling tracks uniformly within each selected grid. We apply this PGR to both perception and controller training since poor performance could come from both modules.
To avoid mode collapse (as observed in~\cite{10.1145/3422622}), we mix a small fraction of uniform sampling $w_i \leftarrow (1-\beta)\,w_i + \beta/M$, and reuse a fixed $G_{\mathrm{val}}$ for stable scoring. 
“Ours (w/ PGR)” in Table~\ref{tab:UAV-result} and Table \ref{tab:quadrotor-result} corresponds to performance-guided refinement algorithm, while “Ours (w/o PGR)” refers to the approach of uniform sampling two-gate tracks.

\section{Experiments}
\label{sec:experiments}

We evaluate our approach in two case studies: fixed-wing UAVs and quadrotors. 
For each case study, we first describe the experimental setups, and then quantitatively evaluate our visual policy against baselines in FalconGym 2.0.
We further demonstrate zero-shot sim-to-real transfer on the quadrotor case study in Section \ref{sec:quadrotor-case-study}. 
%

\subsection{Case Study 1: fixed-wing UAV}
\label{sec:UAV-case-study}

We model the fixed-wing UAV with a Dubins Airplane dynamics~\cite{Owen2015}. 
The state is $(x,y,z,\psi,\theta)$ (3D position, yaw and pitch).
The control action is bank rate and pitch rate.
Forward speed is fixed at $7\,\mathrm{m/s}$. 
We implement the state-based controller also from~\cite{Owen2015}, to serve as a baseline and the expert of our imitation learning approach (Section \ref{sec:vision-controller-architecture}).

Our UAV flying arena measures $40\times20\times4\,\mathrm{m}$.
The square gates used in the UAV case study have an inner side length of $200\,\mathrm{cm}$. 
The UAV has a width of $40\,\mathrm{cm}$.
A gate crossing is deemed \emph{success} if, at the instant the aerial vehicle passes the gate plane, the distance from the UAV to the gate center is less than $80\,\mathrm{cm}$.
We evaluate the performance of gate crossing on two metrics: \emph{Success Rate (SR)},  the percentage of gates the visual policy successfully crosses; \emph{Mean Gate Error (MGE)}, average distance between
the UAV and the gate center at the time of gate crossing.

Using FalconGym 2.0's Edit API, we design three dynamically feasible observable tracks (Figure \ref{fig:uav_trajectory_plot}): 
(i) \emph{Spatial-S}, requiring consecutive sharp turns with altitude changes; 
(ii) \emph{Random}, where UAV largely maintains heading while alternating left/right gates and adjusting height;
(iii) \emph{Moving}, where one gate translates from right to upper-left at $2\,\mathrm{m/s}$. 
For each track, we selected 10 slightly different initial poses to evaluate and report the average SR and MGE in Table \ref{tab:UAV-result}.

\begin{figure}[htbp]
    \centering
    \includegraphics[width=0.7\linewidth]{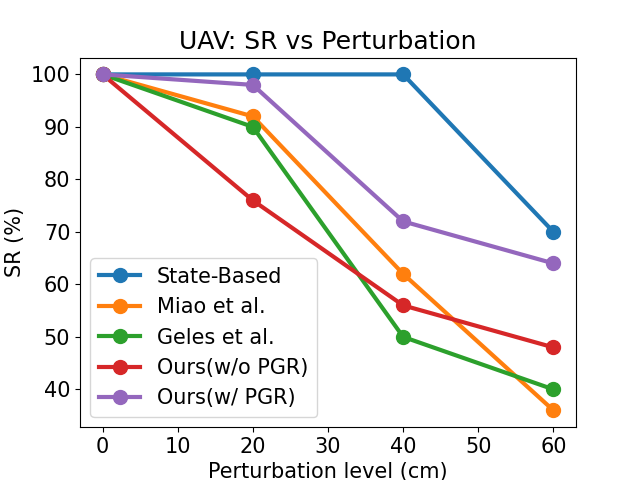}
    \caption{\small{
    \textbf{Robustness to gate-pose perturbations on \emph{Spatial-S} track in FalconGym 2.0.} 
    For a perturbation level $a$\,cm, each gate is independently shifted by a random 3D offset $\delta \in [-a,a]^3$. 
    For each perturbation level, we run all five policies on 10 randomized tracks (50 gates total) and report the Success Rate (SR).
    }}
    \label{fig:UAV-perturbation-plot}
\end{figure}

We implemented five policies and evaluate on their performance in terms of SR, MGE and generalization. 
First, the state-based expert \cite{Owen2015} has access to full state and all gate poses. As expected, it attains the best SR/MGE and generalize well across all tracks.  
Then we implemented two state-of-the-art visual baselines \cite{miao2025zeroshotsimtorealvisualquadrotor, DBLP:conf/rss/GelesBRX024}.
\cite{miao2025zeroshotsimtorealvisualquadrotor} trains the visual policy in a NeRF-based simulation, we adopt their architecture and re-trained the visual policy in our FalconGym 2.0 with the help of our Edit API.
We also replace their IMU input with the past control, because IMU is not applicable for the Dubins airplane model.
\cite{DBLP:conf/rss/GelesBRX024} uses a Swin-transformer-based gate detector, however, since the code is not publicly available, we replaced the gate detector with our Mask Detector describe in Section \ref{sec:vision-controller-architecture}. 
We also modify the reward function by tuning the hyperparameters to make it work in our cases.
We faithfully replicate the controller architecture based on their paper and our produced results, e.g. Figure \ref{fig:UAV-perturbation-plot}, does match their reported behavior.
Since both visual baselines acknowledge in their papers that they require training and testing on the same track, to further evaluate their method's performance on unseen tracks, we train those baselines on the Spatial-S track and evaluate on all three tracks. 
The result shows that while both baselines performs well in the training Spatial-S track, they do not generalize well to unseen tracks Random and Moving, as shown in Table \ref{tab:UAV-result}. 
We also train and test the baselines again on the exact Random track, and find that they can both achieve 100\% success, which confirms that their method are overfitting to training tracks.

\begin{table}[htbp]\centering
\caption{\small{Evaluation of five policies based on Success Rate (SR) and Mean Gate Error (MGE) in the UAV Case Study. Success criteria requires distance at gate plane crossing (MGE) $\leq 80$ cm. }}
\label{tab:UAV-result}
\scriptsize
\resizebox{\columnwidth}{!}{
\begin{tabular}{lccccc}\toprule
      & \multicolumn{5}{c}{FalconGym 2.0} \\ \midrule
    Track & Method & Vision? & Generalize? & SR $\uparrow$ & MGE (cm) $\downarrow$   \\ \midrule
    
    Spatial-S & \underline{State-based} & \xmark & \cmark & \underline{100\%}  & \underline{15} \\
    & Baseline A \cite{miao2025zeroshotsimtorealvisualquadrotor} & \cmark & \xmark & 100\% & 24    \\
    & Baseline B \cite{DBLP:conf/rss/GelesBRX024} & \cmark & \xmark & 100\% & 20    \\
    & Ours (w/o PGR) & \cmark & \cmark & 100\% & 47    \\
    & \cellcolor{gray!50} \textbf{Ours (w/ PGR)} & \cellcolor{gray!50} \cmark & \cellcolor{gray!50} \cmark & \cellcolor{gray!50} \textbf{100\%} &  \cellcolor{gray!50} \textbf{19}     \\\midrule
    
    Random & \underline{State-based} & \xmark & \cmark & \underline{100\%} & \underline{8}     \\
    & Baseline A \cite{miao2025zeroshotsimtorealvisualquadrotor} & \cmark & \xmark & 0\% & N/A    \\
    & Baseline B \cite{DBLP:conf/rss/GelesBRX024} & \cmark & \xmark & 50\% &  47    \\
    & Ours (w/o PGR) & \cmark & \cmark & 100\% &  33    \\
    & \cellcolor{gray!50} \textbf{Ours (w/ PGR)} & \cellcolor{gray!50} \cmark & \cellcolor{gray!50} \cmark & \cellcolor{gray!50} \textbf{100\%} &  \cellcolor{gray!50} \textbf{21} \\ \midrule

    Moving & \underline{State-based} & \xmark & \cmark & \underline{100\%}  & \underline{14}     \\
    & Baseline A \cite{miao2025zeroshotsimtorealvisualquadrotor} & \cmark & \xmark & 0\% & N/A    \\
    & Baseline B \cite{DBLP:conf/rss/GelesBRX024} & \cmark & \xmark & 0\% & N/A    \\
    & Ours (w/o PGR) & \cmark & \cmark & 100\% & 47    \\
    & \cellcolor{gray!50} \textbf{Ours (w/ PGR)} & \cellcolor{gray!50} \cmark & \cellcolor{gray!50} \cmark & \cellcolor{gray!50} \textbf{100\%} & \cellcolor{gray!50} \textbf{15}   \\\midrule
\end{tabular}
}
\end{table}

In contrast, our methods, both our uniformly sampling approach and performance-guided refinement approach generalize to unseen tracks with a \emph{single} visual policy.
In fact, all three evaluation tracks are unseen to our visual policy, since our training only relies on two-gate track layouts (Section \ref{sec:vision-controller-architecture}).
The ``Ours (w/o PGR)'' approach refers to our method by uniformly sampling the gate space to generate $300$k dynamically feasible and observable two-gate tracks and train the visual policy by imitation.
The ``Ours (w/ PGR)'' approach utilized the performance-guided refinement in Section \ref{sec:min-max-optimization}.
Specifically, we partition $G\subset\mathbb{R}^8$ (each gate has $(x,y,z,\text{yaw})$) into $(4\times4\times3\times3)^2\!\approx\!20$k grids. 
In the first iteration we draw $5$ samples per grid to form $100$k training tracks; we then run PGR for $T{=}3$ iterations with $\beta{=}0.05$ in Algorithm~\ref{alg:minmax}. 
Both our variants achieve high SR and generalize well on all tracks, while PGR further reduces MGE, indicating safer crossings and better performance.

Furthermore, we assess each policy's sensitivity to gate position changes using domain randomization enabled by FalconGym 2.0's Edit API, as shown in Figure \ref{fig:UAV-perturbation-plot}. 
For fair comparison, we conduct this ablation study on the Spatial-S track on which both baselines are trained, and evaluate perturbed variants of that track.
At a given perturbation level $a$ (cm), every gate is independently shifted by a 3D offset $\delta \in [-a,a]^3$ (applied along $x$, $y$, and $z$ direction with random signs). 
We generate 10 perturbed tracks (50 gates total) for each perturbation level and run all five policies, reporting Success Rate (SR).
Note that since we originally designed Spatial-S track to be challenging with sharp turns and altitude changes, increasing $a$ demands sharper turns and pitch changes, so SR decreases for all methods. For high perturbation level, even state-based expert fails because of dynamically infeasible tracks.
Figure \ref{fig:UAV-perturbation-plot} shows that our PGR approach remains more robust in gate poses perturbation than the visual baselines and uniform sampling. PGR tracks the expert’s trend closely with only slight SR gap, which is expected since our visual policy imitates state-based expert.

\subsection{Case Study 2: Quadrotor}
\label{sec:quadrotor-case-study}

\subsubsection{Performance in FalconGym 2.0}

\begin{figure}[htbp]
    \centering
    \includegraphics[width=\linewidth]{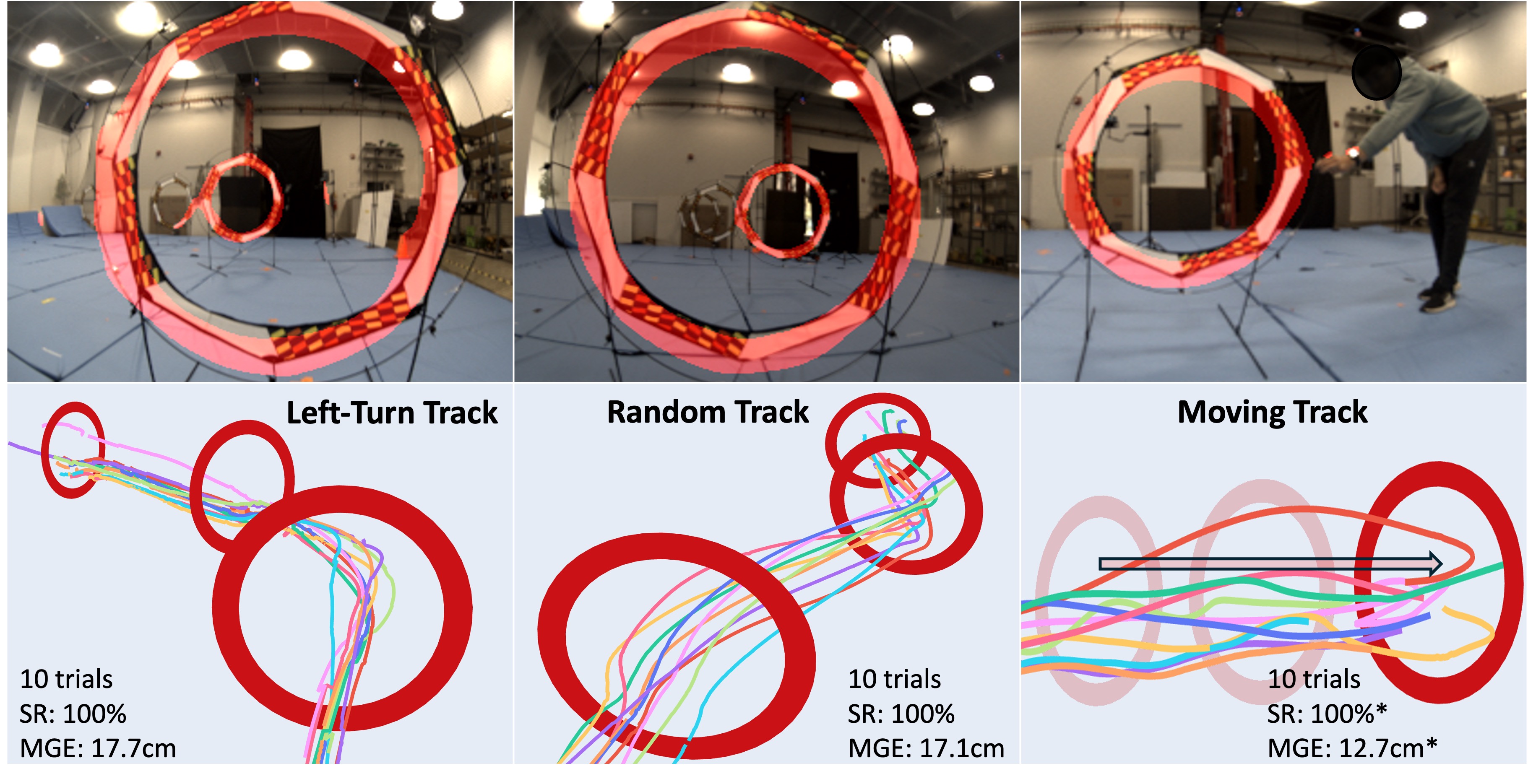}
    \caption{\small{
    \textbf{Trajectory plots for the quadrotor case study on real hardware.} 
    Visual policy learned in FalconGym2.0 can be zero-shot transferred to real hardware. 
    We disable the forward velocity on Moving track for safety reasons, evaluating lateral tracking only.
    }}
    \label{fig:quadrotor_hardware_trajectory_plot}
\end{figure}

We model the quadrotor with the standard 12-state dynamics of~\cite{Sabatino2015QuadrotorCM}. 
The state comprises position, linear velocity, attitude, and angular velocity. 
The state-based expert controller is also from \cite{Sabatino2015QuadrotorCM}, which we set a constant 1m/s forward velocity.
Among several control modalities, we choose body-frame linear velocities and yaw rate to match the hardware interface used for zero-shot sim-to-real transfer in Section \ref{sec: sim2real-transfer}.
Our quadrotor flight arena measures $6\times6\times3\,\mathrm{m}$, as shown in Figure \ref{fig:quadrotor_hardware_trajectory_plot}. 
The circle gates we used have a $78\,\mathrm{cm}$ inner diameter; the quadrotor width is $18\,\mathrm{cm}$.
Therefore, a crossing is considered \emph{successful} if, at the instant the quadrotor intersects the gate plane, its distance to the gate center (MGE) $\leq 30cm$. 
We adapt the same metrics as in the fixed-wing study: \emph{Success Rate (SR)} and \emph{Mean Gate Error (MGE)}.

We again design 3 dynamically feasible and observable tracks (Figure \ref{fig:quadrotor_trajectory_plot}) using FalconGym 2.0's Edit API: 
(i). \emph{Left-Turn}, which involves sustained turning; 
(ii). \emph{Random}, where quadrotors need to maintain general heading while alternating left/right gates;
(iii). \emph{Moving}, where one gate translates to the right at $0.25\,\mathrm{m/s}$.
%

\begin{table}[htbp]
\centering
\caption{\small{Evaluation of five policies based on SR and MGE in both FalconGym 2.0 and real world for the quadrotor case study. * indicates lateral-tracking-only for Moving track for safety reasons. Success criteria requires MGE $\leq 30$ cm.}}
\label{tab:quadrotor-result}
\scriptsize
\resizebox{\columnwidth}{!}{
\begin{tabular}{lccc|cc}\toprule
    &  & \multicolumn{2}{c|}{FalconGym 2.0} & \multicolumn{2}{c}{Real World} \\\midrule
    Track & Method & SR $\uparrow$ & MGE (cm) $\downarrow$ & SR $\uparrow$ & MGE (cm) $\downarrow$ \\\midrule
    
    Left-Turn & \underline{State-based} & \underline{100\%}  & \underline{5.8}  & \underline{100\%}  &  \underline{11.6}\\
    & Baseline A \cite{miao2025zeroshotsimtorealvisualquadrotor} & 100\% & 14.3 & / & / \\
    & Baseline B \cite{DBLP:conf/rss/GelesBRX024} & 100\% & 17.0 &  / & / \\
    & Ours (w/o PGR) & 100\% &  11.3 & 93.3\% & 21.4 \\
    & \cellcolor{gray!50} \textbf{Ours (w/ PGR)} & \cellcolor{gray!50} \textbf{100\%} & \cellcolor{gray!50} \textbf{9.5} & \cellcolor{gray!50} \textbf{96.7\%} & \cellcolor{gray!50} \textbf{17.7} \\\midrule
    
    Random & \underline{State-based} & \underline{100\%}  & \underline{5.7}  & \underline{100\%} & \underline{5.9} \\
    & Baseline A \cite{miao2025zeroshotsimtorealvisualquadrotor} &  67.7\% &  32.3 & / & / \\
    & Baseline B \cite{DBLP:conf/rss/GelesBRX024} & 25\% & 40.0 & / & / \\
    & Ours (w/o PGR) & 100\% & 11.4 & 96.7\% & 16.2 \\
    & \cellcolor{gray!50} \textbf{Ours (w/ PGR)} & \cellcolor{gray!50} \textbf{100\%} & \cellcolor{gray!50} \textbf{9.5} & \cellcolor{gray!50} \textbf{100\%} & \cellcolor{gray!50} \textbf{17.1} \\\midrule

    Moving & \underline{State-based} & \underline{100\%} & \underline{2.0} & \underline{100\%*} & \underline{7.4*} \\
    & Baseline A \cite{miao2025zeroshotsimtorealvisualquadrotor} & 0\% & N/A & / & / \\
    & Baseline B \cite{DBLP:conf/rss/GelesBRX024} & 0\% & N/A & / & / \\
    & Ours (w/o PGR) & 100\% & 24.1 & 100\%*  & 16.8* \\
    & \cellcolor{gray!50} \textbf{Ours (w/ PGR)} & \cellcolor{gray!50} \textbf{100\%} & \cellcolor{gray!50} \textbf{20.7} & \cellcolor{gray!50} \textbf{100\%*}  & \cellcolor{gray!50} \textbf{12.7*} \\\midrule
\end{tabular}
}
\end{table}

Baselines are implemented as in the fixed-wing study but trained with the quadrotor GSplat data, with the modifications described in Section~\ref{sec:UAV-case-study} to fit our pipeline. 
While our exact visual policy (perception and controller) also works in the quadrotor case study, we slightly modify it by replacing the imitation-learned controller with a classical controller that (i) filters the perception noise and retains only the largest connected component (closest gate) in the predicted mask and (ii) commands the aerial vehicle to follow the centroid of that component. 
This engineering trade-off is necessary because the hardware quadrotor (Section \ref{sec: sim2real-transfer}) cannot execute two neural networks sequentially at the required closed-loop rate. 
Therefore, we keep the U-Net for perception and replace the second network with a classical controller. 
As a result, instead of applying PGR to the entire visual policy (perception and controller) as in the fixed-wing UAV case study, we apply it only to the perception module.

As in the fixed-wing UAV experiments, we evaluate all five policies on three tracks with ten different initial conditions. 
Again, both baselines perform well on the training \emph{Left-Turn} track but overfit and degrade on the unseen \emph{Random} and \emph{Moving} tracks, as shown in Table~\ref{tab:quadrotor-result}. 
By contrast, both our uniform sampling (Ours w/o PGR) and performance-guided refinement (Ours w/ PGR) maintain $100\%$ SR across all tracks. 
While PGR further improves MGE, the margin over uniform sampling is smaller than in the fixed-wing case. 
We believe this is due to: (i) PGR is only applied to the perception module (rather than the full perception-control stack) and (ii) the quadrotor’s smaller workspace with coarser grid discretization, i.e., $\bigl(3\times3\times2\times2\bigr)^2 = 1{,}296$ grids.

\begin{figure}[htbp]
    \centering
    \includegraphics[width=0.7\linewidth]{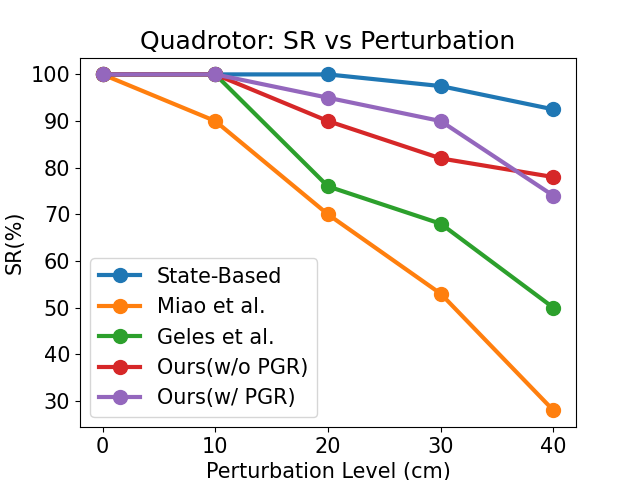}
    \caption{\small{
    \textbf{Gate perturbations on Left-Turn track in FalconGym 2.0.} 
    For each perturbation level, we run all five policies on 10 randomized tracks (30 gates total) and report the Success Rate (SR).
    }}
    \label{fig:quadrotor-perturbation-plot}
\end{figure}

We also conduct a gate-pose perturbation study similar to the fixed-wing ablation study on the Left-Turn track in FalconGym 2.0.
Figure~\ref{fig:quadrotor-perturbation-plot} shows that baselines' SR drops rapidly as perturbation level increases, whereas our policies remain substantially more robust and closely track the expert’s.
The incremental gain from PGR over uniform sampling is modest, again likely due to the partial refinement on the perception only and the smaller space and number of grids.


\subsubsection{Zero-shot Sim-to-Real Transfer}
\label{sec: sim2real-transfer}

We zero-shot deployed the trained perception module and same classical controller used in FalconGym 2.0 on a 280\,g ModalAI\textsuperscript{\textregistered} Starling~2 quadrotor (upper-right of Figure~\ref{fig:closed-loop}). 
The quadrotor is equipped with a PX4 flight controller, a VOXL~2\textsuperscript{\textregistered} onboard computer with an Adreno 650 GPU for neural inference, and a 12\,MP RGB camera. 
The onboard closed-loop frequency is 8\,Hz when running U-Net perception and classical controller.
%

Although our visual policy does not require the tracks to be the same as in training or in FalconGym 2.0, for a fair sim-to-real comparison, we arranged the hardware courses to mirror the simulation layouts: \emph{Left-Turn}, \emph{Random}, and \emph{Moving}. 
Due to the small indoor arena and observability assumptions, for \emph{Left-Turn} and \emph{Random} we cross only three gates and reserve the fourth as a navigational waypoint to satisfy the observability requirement. 
For \emph{Moving} track, a human operator slowly pulls the gate laterally at around \( 0.25\,\mathrm{m/s}\); we disable forward velocity for safety reasons and evaluate average lateral tracking error of the gate centroid as MGE. 
We use a motion-capture (mocap) to log the quadrotor's poses and the moving gate's trajectories.
While we also use mocap to implement the state-based controller baseline, it is not used in FalconGym 2.0's construction or in any of the vision-based closed-loop.
Real-world onboard views and quadrotor trajectories are visualized in Figure~\ref{fig:quadrotor_hardware_trajectory_plot}. 

Due to safety concerns, differing control modalities, and limited onboard computation power, we do not run the visual baselines on hardware. 
We evaluate the remaining three policies with the same settings in FalconGym 2.0: each policy is run on each track with 10 different initial conditions, and then we report the SR and MGE in Table \ref{tab:quadrotor-result}. 
Although both our visual-policy variants perform slightly worse than in simulation, they maintain high success rates (SR \(\geq 93\%\)) across all tracks. 
The primary failure modes are (i) perception errors where the U-Net confuses a background ladder with a gate and (ii) latency limits from the 8\,Hz closed loop. 
We mark the Moving track results with * to indicate lateral-tracking-only evaluation in hardware (no gate crossing due to safety reasons), which most likely causes lower MGE than the full crossing achieved in FalconGym 2.0. 
Consistent with simulation and the UAV case study, PGR shows a slight improvement over uniform sampling in SR and MGE. 
And both our visual policy variants follow the state-based controller performance closely in 3 tracks.

\section{CONCLUSIONS}

We introduced \emph{FalconGym~2.0} (code to be released upon publication), a GSplat-based photorealistic simulation framework that provides an \emph{Edit API} for programmatic object transforms. 
Leveraging this editability, we proposed a \emph{performance-guided refinement} (PGR) algorithm that concentrates visual policy's training on challenging tracks and iteratively improves its performance. 
Across two case studies (fixed-wing UAVs and quadrotors) with different dynamics and environments, our visual policy generalizes to different tracks, outperforms baselines by achieving $100\%$ SR on unseen tracks in FalconGym~2.0 and showing better robustness to gate-pose perturbations. 
Finally, we demonstrate zero-shot sim-to-real transfer on a hardware quadrotor with $98.6\%$ SR over three tracks (70 gates).

In future work, we plan to: relax the \emph{observability} assumption to handle occluded gates; evaluate higher-speed flights for the quadrotor case study; incorporate realistic UAV dynamics to enable fixed-wing sim-to-real transfer; and distill the perception and controller neural networks to deploy the full two-network stack onboard.









\bibliographystyle{IEEEtran}
\bibliography{reference}

\end{document}